\pdfoutput=1
\documentclass[11pt,a4paper]{article}
\usepackage[hyperref]{emnlp2020}
\usepackage{times}
\usepackage{latexsym}

\usepackage{booktabs}
\usepackage{amsmath}
\usepackage{stfloats}
\usepackage{multirow,hhline,graphicx,array}
\usepackage{url}
\newcommand\Tstrut{\rule{0pt}{2.6ex}}
\newcommand\Bstrut{\rule[-0.9ex]{0pt}{0pt}}

\usepackage{microtype}

\aclfinalcopy 

\title{What do Models Learn from Question Answering Datasets?}

\author{Priyanka Sen \\
  Amazon Alexa\\
  \texttt{sepriyan@amazon.com} \\\And
  Amir Saffari \\
  Amazon Alexa\\
  \texttt{amsafari@amazon.com} \\}

\date{}

\begin{document}
\maketitle
\begin{abstract}
While models have reached superhuman performance on popular question answering (QA) datasets such as SQuAD, they have yet to outperform humans on the task of question answering itself. In this paper, we investigate if models are learning reading comprehension from QA datasets by evaluating BERT-based models across five datasets. We evaluate models on their generalizability to out-of-domain examples, responses to missing or incorrect data, and ability to handle question variations. We find that no single dataset is robust to all of our experiments and identify shortcomings in both datasets and evaluation methods. Following our analysis, we make recommendations for building future QA datasets that better evaluate the task of question answering through reading comprehension. We also release  code to convert QA datasets to a shared format for easier experimentation at \url{https://github.com/amazon-research/qa-dataset-converter}.
\end{abstract}

\section{Introduction}
Question answering (QA) through reading comprehension has seen considerable progress in recent years. This progress is in large part due to the release of large language models like BERT \cite{devlin-etal-2019-bert} and new datasets that have introduced impossible questions \cite{rajpurkar-etal-2018-know}, bigger scales \cite{kwiatkowski-etal-2019-natural}, and context \cite{choi-etal-2018-quac, reddy-etal-2019-coqa} to question answering. At the time of writing this paper, models have outperformed human baselines on the widely-used SQuAD 1.1 and SQuAD 2.0 datasets, and more challenging datasets such as QuAC have models less than 7 F1 points away. Despite these increases in F1 scores, we are still far from saying question answering is a solved problem. 

Concerns have been raised over how challenging QA datasets are. Previous work has found that simple heuristics can give good performance on SQuAD 1.1 \cite{weissenborn-etal-2017-making}, and successful SQuAD models lack robustness by giving inconsistent answers \cite{ribeiro-etal-2019-red} or being vulnerable to adversarial attacks \cite{jia-liang-2017-adversarial, wallace-etal-2019-universal}. If state-of-the-art models are excelling at test sets but not solving the underlying task of reading comprehension, then our test sets are flawed. We need to better understand what models learn from QA datasets. In this work, we ask three questions: (1) Does performance on individual QA datasets generalize to new datasets? (2) Do models need to learn reading comprehension for QA datasets?, and (3) Can QA models handle variations in questions?

To answer these questions, we evaluate BERT models trained on five QA datasets using simple generalization and robustness probes. We find that (1) Model performance does not generalize well outside of  heuristics like question-context overlaps, (2) Removing or corrupting dataset examples does not always harm model performance, showing that models can rely on simpler methods than reading comprehension to answer questions, and (3) No dataset fully prepares models to handle question variations like filler words or negation. These findings suggest that while QA models can learn heuristics around question-context overlaps and named entities, they do not need to learn reading comprehension to perform well on QA datasets. Based on these findings, we make recommendations on how to better create and evaluate QA datasets.

\section{Related Work}
Our work is inspired by recent trends in NLP to evaluate generalizability and probe what models learn from datasets. In terms of generalizability, prior work has been done by \citet{yogatama2019learning} who evaluated a SQuAD 1.1 model across four datasets and \citet{talmor-berant-2019-multiqa}, who comprehensively evaluated ten QA datasets. The MRQA 2019 shared task \cite{fisch2019mrqa} evaluated transferability across multiple datasets, and \newcite{khashabi2020unifiedqa} proposed a method to train one model on 17 different QA datasets. In our work, we focus on question answering through reading comprehension and extend the work on generalizability by including impossible questions in all our datasets and analyzing the effects of question-context overlap on generalizability.

There is also growing interest in probing what models learn from datasets \cite{mccoy-etal-2019-right,geirhos2020shortcut, richardson2020does,si2020benchmarking}. Previous work in question answering has found that state-of-the-art models can get good performance on incomplete input \cite{agrawal-etal-2016-analyzing, sugawara-etal-2018-makes, niven-kao-2019-probing}, under-rely on important words, \cite{mudrakarta-etal-2018-model}, and over-rely on simple heuristics \cite{weissenborn-etal-2017-making,ko2020look}. Experiments on SQuAD in particular have shown that SQuAD models are vulnerable to adversarial attacks \cite{jia-liang-2017-adversarial, wallace-etal-2019-universal} and not robust to paraphrases \cite{ribeiro-etal-2018-semantically, gan-ng-2019-improving}. 

Our work continues exploring what models learn by comprehensively testing multiple QA datasets against a variety of simple but informative probes. We take inspiration from previous studies, and we make novel contributions by using BERT, a state-of-the-art model, and running several experiments against five different QA datasets to investigate the progress made in reading comprehension.

\section{Datasets}

\begin{table}[htb]
\begin{center}
\begin{tabular}{l c c}
\toprule
& Train & Dev \\
\midrule
SQuAD & 130,319 & 11,873 \\
TriviaQA & 110,647 & 14,229 \\
NQ & 110,857 & 3,368 \\
QuAC & 83,568 & 7,354 \\
NewsQA & 101,707 & 5,666 \\
\bottomrule
\end{tabular}
\end{center}
\caption{Train and dev set sizes of the datasets used in our experiments}
\label{tab:datasets_splits}
\end{table}

\begin{table}[htb]
\begin{center}
\begin{tabular}{l c c c}
\toprule
& Question & Context & Answer \\
\midrule
SQuAD & 10 & 120 & 3 \\
TriviaQA & 15 & 746 & 2 \\
NQ & 9 & 96 & 4 \\
QuAC & 7 & 395 & 14 \\
NewsQA & 8 & 709 & 4 \\
\bottomrule
\end{tabular}
\end{center}
\caption{Comparison of the average number of words in questions, contexts, and answers in each dataset}
\label{tab:datasets}
\end{table}

We compare five datasets in our experiments: SQuAD 2.0, TriviaQA, Natural Questions, QuAC, and NewsQA. All our datasets treat question answering as a reading comprehension task where the question is about a document and the answer is either an extracted span of text or labeled unanswerable. To consistently compare and experiment across models, we convert all datasets into a SQuAD 2.0 JSON format.\footnote{\url{https://github.com/amazon-research/qa-dataset-converter}} Since most datasets have a hidden test set, we evaluate models on the dev set and consequently refer to the dev sets as test sets in this paper. The train and dev sets sizes are shown in Table \ref{tab:datasets_splits}

Below we describe each dataset and any modifications we made to run our experiments:

\textbf{SQuAD 2.0} \cite{rajpurkar-etal-2018-know} consists of 150K question-answer pairs on Wikipedia articles. To create SQuAD 1.1, crowd workers wrote questions about a Wikipedia paragraph and highlighted the answer \cite{rajpurkar-etal-2016-squad}. SQuAD 2.0 includes an additional 50K plausible but unanswerable questions written by crowd workers.
  
\textbf{TriviaQA} \cite{joshi-etal-2017-triviaqa} includes 95K question-answer pairs from trivia websites. The questions were written by trivia enthusiasts and the evidence documents were retrieved by the authors retrospectively. We use the variant of TriviaQA where the documents are Wikipedia articles.
  
\textbf{Natural Questions (NQ)} \cite{kwiatkowski-etal-2019-natural} contains 300K questions from Google search logs. For each question, a crowd worker found a long and short answer on a Wikipedia page. We use the subset of NQ with a long answer and frame the task as finding the short answer in the long answer. We only include examples with answers in the paragraph text (as opposed to a table or list).
  
\textbf{QuAC} \cite{choi-etal-2018-quac} contains 100K questions. To create QuAC, one crowd worker asked questions about a Wikipedia article to a second crowd worker, who answered by selecting a text span. To standardize training, we do not model contextual information, but we include QuAC to see how models trained without context handle context-dependent questions.
  
\textbf{NewsQA} \cite{trischler-etal-2017-newsqa} contains 100K questions on 10K CNN articles. One set of crowd workers wrote questions based on a headline and summary, and a second set of workers found the answer in the article. We reintroduce unanswerable questions that were excluded in the original paper.

There are notable differences among our datasets in terms of genre and how they were built. In Table \ref{tab:datasets}, we see a large variation in the average number of words in questions, contexts, and answers. Despite these differences, all our datasets are reading comprehension tasks. We believe a good reading comprehension model should handle question answering well regardless of dataset differences, and so we compare across all five datasets. 

\begin{table*}[ht]
\centering
\begin{tabular}{c l c c c c c}
\toprule
 \multicolumn{7}{c}{Evaluated on} \\
 & & SQuAD & TriviaQA & NQ & QuAC & NewsQA\Tstrut\Bstrut\\
 \cline{2-7}
 & SQuAD & \textbf{75.6} & 46.7 & 48.7 & 20.2 & 41.1\Tstrut\Bstrut\\
 & TriviaQA & 49.8 & \textbf{58.7} & 42.1 & 20.4 & 10.5\Tstrut\Bstrut\\
\raisebox{0\normalbaselineskip}[0pt][0pt]{\rotatebox[origin=c]{90}{Fine-tuned on}} & NQ & 53.5 & 46.3 & \textbf{73.5} & 21.6 & 24.7\Tstrut\Bstrut\\
 & QuAC & 39.4 & 33.1 & 33.8 & \textbf{33.3} & 13.8 \Tstrut\Bstrut\\
 & NewsQA & 52.1 & 38.4 & 41.7& 20.4 & \textbf{60.1} \Tstrut\Bstrut\\
\bottomrule
\end{tabular}
\caption{F1 scores of each fine-tuned model evaluated on each test set}
\label{tab:eval}
\end{table*}

\section{Model}

\begin{table}[hbt!]
\begin{center}
\begin{tabular}{l c}
\toprule
Hyperparameter & Value \\
\midrule
Batch Size & 24\\
Learning Rate & 3e-5\\
Epochs & 2\\
Max Seq Length & 384\\
Doc Stride & 128\\
\bottomrule
\end{tabular}
\end{center}
\caption{Hyperparameter values for fine-tuning BERT based on \citet{devlin-etal-2019-bert}}
\label{tab:hyperparam}
\end{table}

All models are initialized from a pre-trained BERT-Base uncased model\footnote{\url{https://github.com/google-research/bert\#pre-trained-models}} with 110M parameters. For each dataset, we fine-tune on the training set using \citet{devlin-etal-2019-bert}'s default hyperparameters shown in Table~\ref{tab:hyperparam}. We evaluate on the dev set with the SQuAD 2.0 evaluation script \cite{rajpurkar-etal-2018-know}. We run our experiments on a single Nvidia Tesla v100 16GB GPU.

In Table~\ref{tab:comparison}, we provide a comparison between our models and previously published BERT results. Differences occur when we make modifications to match SQuAD. We simplified NQ by removing the long answer identification task and framed the short answer task in a SQuAD format, so we see higher results than the NQ BERT baseline. For QuAC, we ignored all context-related fields and treated each example as an independent question, so we see lower results than models built on the full dataset. For NewsQA, we introduced impossible questions, resulting in lower performance. We accept these drops in performance since we are interested in comparing changes to a baseline rather than achieving state-of-the-art results.

\begin{table}[htb]
\begin{center}
\begin{tabular}{l l l}
\toprule
Dataset & Reference & Ours \\
\midrule
SQuAD & 76.3 \cite{liu2019roberta} & 75.6 \\
TriviaQA & 56.3 \cite{yang2019data} & 58.7 \\
NQ & 52.7 \cite{alberti2019bert} & 73.5 \\
QuAC & 54.4 \cite{qu2019bert} & 33.3 \\
NewsQA & 66.8 \cite{takahashi-etal-2019-cler} & 60.1 \\
\bottomrule
\end{tabular}
\end{center}
\caption{Comparison to previously reported F1 scores. Differences occur when we make modifications to match SQuAD.}
\label{tab:comparison}
\end{table}

\section{Experiments}
In this section, we discuss the experiments run to evaluate what models learn from QA datasets. All results are reported as F1 scores since they are correlated with Exact Match scores and are more forgiving to sometimes arbitrary cutoffs of answers (for example, we prefer to give some credit to a model for selecting “Charles III” even if the answer was “King Charles III”).

\subsection{Does performance on individual QA datasets generalize to new datasets?}

\begin{figure}
\centering
\includegraphics[width=0.48\textwidth, clip]{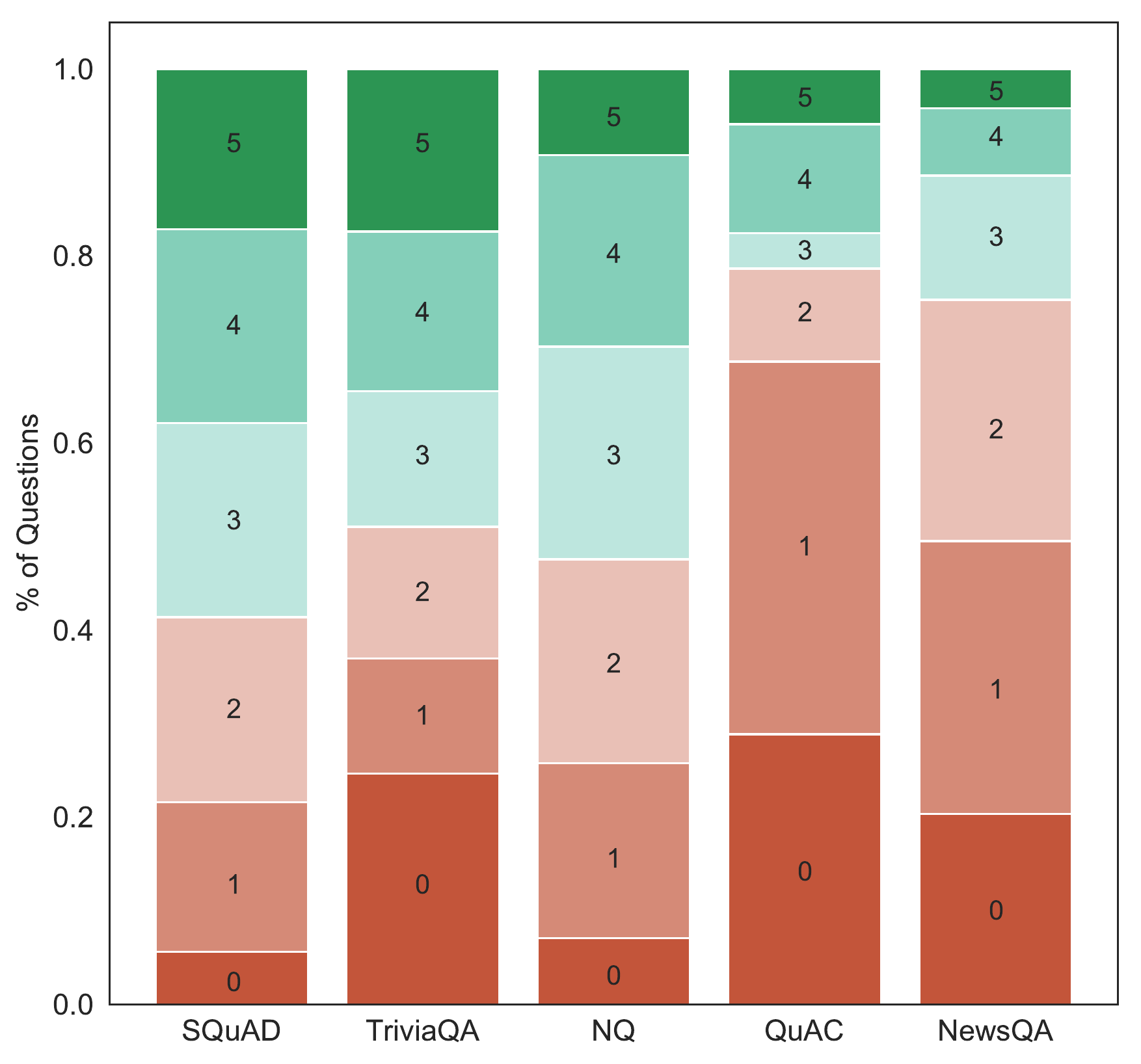}
\caption{A bar graph of how many questions in each dataset are answered by 0, 1, 2, 3, 4, or 5 models}
\label{fig:dataset-diff}
\end{figure}

For our first experiment, we evaluate the generalizability of models to out-of-domain examples. While most work in QA has focused on evaluating against a single test set, generalizability is an important feature. If we cannot get good, generalizable performance on research datasets, we will struggle to expand to the variety of questions an open-domain QA system can face. Several papers have focused on generalizability by evaluating transferability across datasets \cite{talmor-berant-2019-multiqa, yatskar-2019-qualitative}, generalizability to out-of-domain data \cite{fisch2019mrqa}, and building cross-dataset evaluation methods \cite{dua2019orb}.

We test generalizability by fine-tuning models on each dataset and evaluating against all five test sets. The results are reported as F1 scores in Table~\ref{tab:eval}. The rows show a single model’s performance across all five datasets, and the columns show the performance of all the models on a single test set. The model-on-self baseline is indicated in bold.

All models take a drop in performance when evaluated on an out-of-domain test set. This shows that performance on an individual dataset does not generalize across datasets, confirming results found on different mixes of datasets \cite{talmor-berant-2019-multiqa, yogatama2019learning}. However there is variation in how the models perform. For example, models score up to 53.5 F1 points on SQuAD without seeing SQuAD examples, while models do not score above 21.6 F1 points on QuAC without QuAC examples. This suggests that some test sets are easier than others. 

To quantify this, we calculate what proportion of each test set can be correctly answered by how many models. This data is represented as a bar graph in Figure~\ref{fig:dataset-diff}. Each bar represents one dataset, and the segments show how much of the test set is answered correctly by 0 to 5 of the models. 

We consider questions easier if more models correctly answer them. The figure shows that QuAC and NewsQA are more challenging test sets and contain a higher proportion of questions that are answered by 0 or 1 model. In contrast, more than half of SQuAD and NQ and almost half of TriviaQA can be answered correctly by 3 or more models. 

\begin{figure}
\centering
\includegraphics[width=0.48\textwidth, clip]{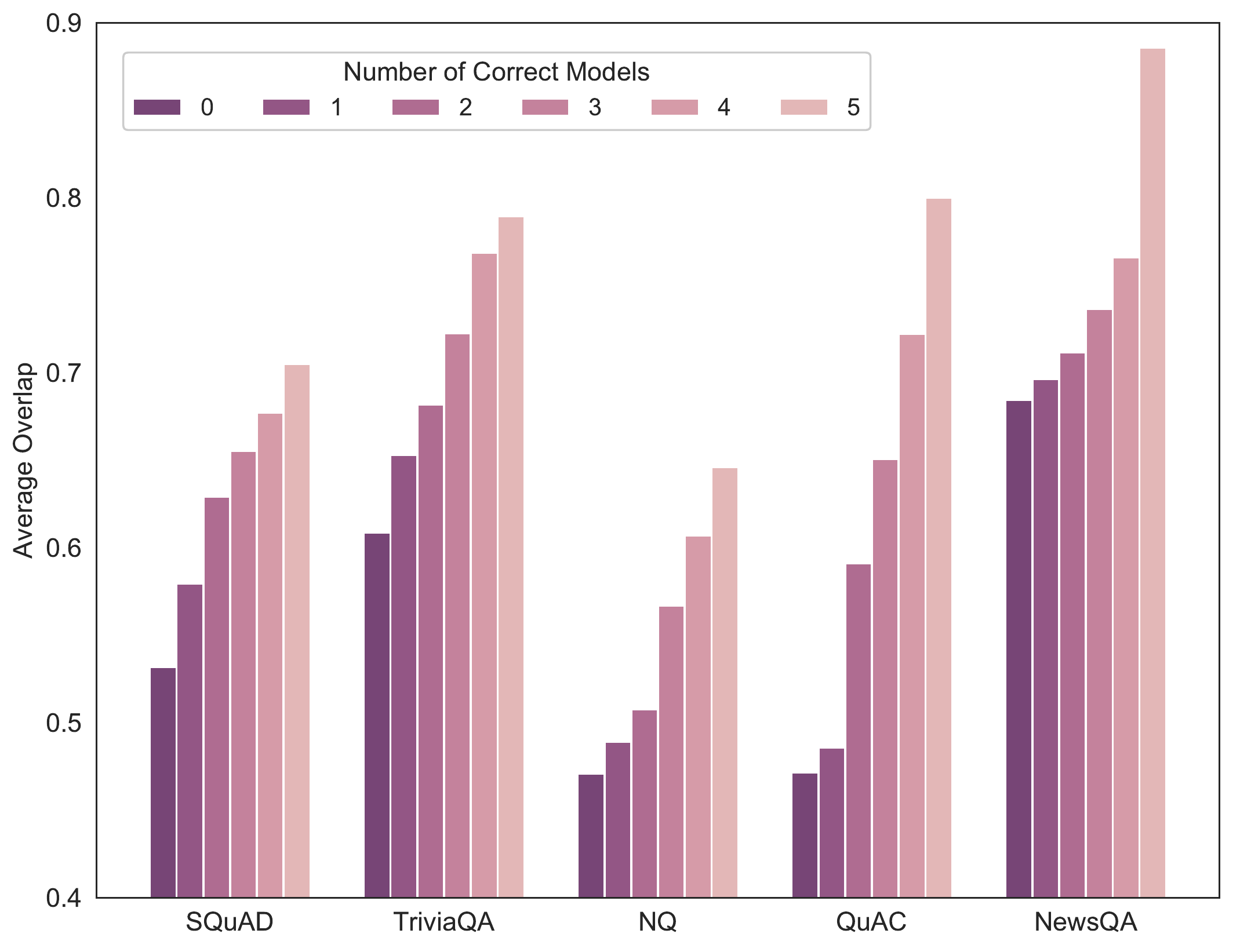}
\caption{More models correctly answer answerable questions if they have higher question-context overlap.}
\label{fig:overlap}
\end{figure}

\begin{figure}
\centering
\includegraphics[width=0.48\textwidth, clip]{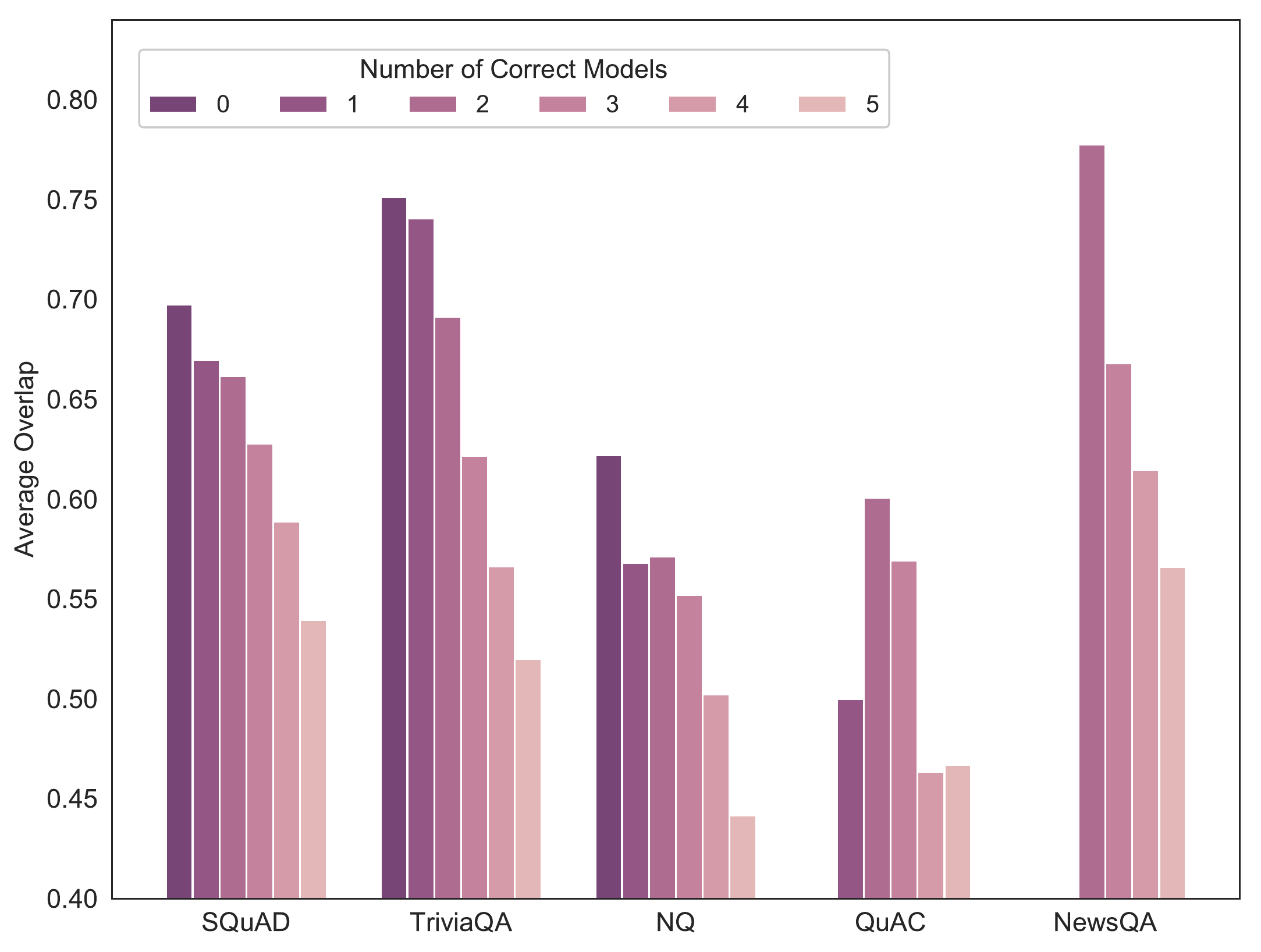}
\caption{More models correctly answer impossible questions if they have lower question-context overlap. NewsQA has four bars since all impossible NewsQA questions were correctly answered by at least 1 model.}
\label{fig:overlap-impossible}
\end{figure}

While difficult questions pose a challenge for QA models, too many easy questions inflate our understanding of a model’s performance. What makes a question easy? We identified a trend between question difficulty and the overlap between the question and the context. We measured overlap as the number of words that appeared in both the question and the context divided by the number of words in the question. For answerable questions, Figure~\ref{fig:overlap} shows that more models return correct answers when there is higher overlap, while Figure~\ref{fig:overlap-impossible} shows that fewer models correctly identify impossible questions when there is higher overlap. This suggests that models learn to use question-context overlap to identify answers. Models may even over-rely on this strategy and return answers to impossible questions when no answer exists.

Overall, the results show that our models do not generalize well to different datasets. The models do seem to exploit question-context overlap, even on questions that are out-of-domain. Reducing the number of high overlap questions in a dataset could create more challenging datasets in the future and better evaluate generalization and test more complex strategies for question answering.

\begin{table*}[ht]
\begin{center}
\begin{tabular}{l l l c c}
\toprule
& Experiment & Question & Answer Text & Answer Start\\
\midrule
1 & Original & Who was the Norse leader & Rollo & 308 \\
2 & Random Label & Who was the Norse leader & succeeding
& 721\\
3 & Shuffled Context & Who was the Norse leader & Rollo
& 480 \\
4 & Incomplete (first half) & Who was & Rollo
& 308 \\
5 & Incomplete (first word) & Who & Rollo
& 308 \\
6 & Incomplete (no words) & & Rollo
& 308 \\
7 & Filler word & Who really was the Norse leader & Rollo
& 308 \\
8 & Negation & Who wasn't the Norse leader & Rollo
& 308 \\
\toprule
\end{tabular}
\end{center}
\caption{Examples of how question-answer pairs are modified in each experiment}
\label{tab:examples}
\end{table*}

\subsection{Do models need to learn reading comprehension for QA datasets?}

State-of-the-art models get good performance on QA datasets, but does good performance mean good reading comprehension? Or are models able to take shortcuts to arrive at the same answers? We explore this by performing three dataset ablation experiments with random labels, shuffled contexts, and incomplete questions. If models can answer test set questions with incorrect or missing information, then the models are likely not learning the task of reading comprehension. The three experiments and their results are discussed in the next sections.

\subsubsection{Random Labels}

\begin{table}[ht]
\begin{center}
\begin{tabular}{l c c c c}
\toprule
\multicolumn{1}{l}{} &
\multicolumn{1}{l}{} &
\multicolumn{3}{c}{\% of Random Labels}    \\ 
\cmidrule(lr){3-5}
Dataset & Baseline & 10\% & 50\% & 90\% \\
\midrule
SQuAD & 78.5 & 77.1 & 73.9 & 32.1 \\
TriviaQA & 46.8 & 36.6 & 10.9 & 0.0 \\
NQ & 70.6 & 68.1 & 60.5& 19.4 \\
QuAC & 20.3 & 16.4 & 1.8 & 0.3 \\
NewsQA & 56.3 & 50.8 & 30.2 & 2.0 \\
\bottomrule
\end{tabular}
\end{center}
\caption{F1 scores of answered questions decrease as models are fine-tuned on increasingly noisy data.}
\label{tab:random-labels}
\end{table}

A robust model should withstand some amount of noise at training time to offset annotation error. However if a model can perform well with a high level of noise, we should be wary of what the model has learned. In our first dataset ablation experiment, we evaluated how various amounts of noise at training time affected model performance.

To introduce noise to the training sets, we randomly selected 10\%, 50\%, or 90\% of the training examples that were answerable and updated the answer to a random string from the same context and of the same length as the original answer. We ensured that the random answer contained no overlaps with the original answer. For simplicity, we did not alter impossible examples. An example of a random label is in the second row of Table~\ref{tab:examples}.

We fine-tuned new models on increasingly noisy training sets and evaluated them on the original test sets. The results are in Table~\ref{tab:random-labels} in terms of F1 scores and reported only for answerable questions. On training sets with 10\% random labels, all models see an F1 score drop. SQuAD, NQ, and NewsQA achieve over 90\% of their baseline score, showing robustness to a reasonable level of noise. TriviaQA and QuAC take larger F1 hits (achieving only 78\% and 81\% of their baselines), suggesting that they are less robust to this noise.

As the amount of noise increases, most F1 scores drop to nearly 0. SQuAD and NQ, however, are suspiciously robust even when 90\% of their training examples are random. SQuAD achieves 41\% of its baseline and NQ achieves 27\% of its baseline with training sets that are 90\% noise. We find it unlikely that randomly selected strings provide a signal, so this suggests that some examples in each test set are answerable trivially and without learning reading comprehension.

\subsubsection{Shuffled Context}

\begin{table}[ht]
\begin{center}
\begin{tabular}{l c c}
\toprule
Dataset & Baseline & Shuffled Context \\
\midrule
SQuAD & 75.6 & 70.5 \\
TriviaQA & 58.7 & 38.7 \\
NQ & 73.5 & 64.5 \\
QuAC & 33.3 & 32.4 \\
NewsQA & 60.1 & 48.2 \\
\bottomrule
\end{tabular}
\end{center}
\caption{F1 scores decrease, but not dramatically, on test sets with shuffled context sentences.}
\label{tab:shuffled}
\end{table}

The task of reading comprehension aims to measure how well a model understands a given passage. If a model is able to answer questions without understanding the logic or structure of a passage, we can get high scores on a test set but be no closer to learning reading comprehension. In our second experiment, we investigate how much of model performance can be accounted for without understanding the full passage.

For each context paragraph in the test set, we split the context by sentence, randomly shuffled the sentences, and rejoined the sentences into a new paragraph. The original answer text remained the same, but the answer start token was updated by locating the correct answer text in the shuffled context. An example is in the third row of Table~\ref{tab:examples}.

We used our models fine-tuned on the original training sets and evaluated on the shuffled context test sets. The results are in Table~\ref{tab:shuffled}. TriviaQA sees the largest drop in performance, achieving only 66\% of its baseline, followed by NewsQA with 80\% of its baseline. SQuAD and QuAC, on the other hand, get over 93\% of their original baselines even with randomly shuffled contexts. TriviaQA and NewsQA have longer contexts, with an average of over 700 words, and so shuffling longer contexts seems more detrimental. While these results show that models do not rely on naive approaches, like position, they do show that for many questions, models do not need to understand a paragraph's structure to correctly predict the answer. 

\subsubsection{Incomplete Input}

\begin{table}[htb]
\begin{center}
\begin{tabular}{l c c c c}
\toprule
& & First & First & No\\
Dataset & Baseline & Half & Word & Words\\
\midrule
SQuAD & 75.6 & 36.4 & 22.8 & 49.5 \\
TriviaQA & 58.7 & 45.8 & 31.8 & 30.4 \\
NQ & 73.5 & 61.4 & 50.3 & 32.7\\
QuAC & 33.3 & 25.2 & 22.4 & 20.2\\
NewsQA & 60.1 & 43.6 & 26.3 & 13.4 \\
\toprule
\end{tabular}
\end{center}
\caption{F1 scores decrease with incomplete input, but models can return correct answers with no question.}
\label{tab:incomplete}
\end{table}

QA dataset creators and their crowd workers spend considerable effort hand-crafting questions that are meant to challenge a model’s ability to understand language. But are models using the questions? In previous work, \citet{agrawal-etal-2016-analyzing} found that a Visual Question Answering (VQA) model could get good performance with just half the original question, while \citet{sugawara-etal-2018-makes} saw drops in BiDAF model performance on QA datasets with increasingly incomplete questions. We expand on these previous works by using BERT, including impossible questions, and introducing NER baselines for comparison.

We created three variants of each test set containing only the first half, first word, or no words from each question. The answer expectations were not changed. Examples are in the fourth, fifth, and sixth rows of Table~\ref{tab:examples}.

We evaluated models fine-tuned on the original training set on the incomplete test sets. The results are in Table~\ref{tab:incomplete}. F1 scores mostly decrease on test sets with incomplete input, but models can return correct answers without being given the question. SQuAD achieves 65\% of its baseline given no question, an increase from the First Word test set primarily because of higher success on impossible questions. NQ achieves up to 68\% of its baseline F1 score given the first word and up to 44\% given no question. These results show that not all examples require full or any question understanding to make correct predictions. We also see higher F1 scores compared to \citet{sugawara-etal-2018-makes} when using the first word. In TriviaQA, \citet{sugawara-etal-2018-makes} saw their F1 score drop by 75\% (from 49.3 to 12.5) while we see a drop of 46\% (from 58.7 to 31.8), which could suggest that our BERT models have overfit more than BiDAF models.

How can models answer without the full question? We investigated our results further by creating two naive named entity recognition (NER) baselines using spaCy\footnote{\url{https://spacy.io}} to see if models can rely on entity types. For our First Word NER baseline, we used the first word of the question to choose an entity as the answer. If a question started with “who”, we returned the first person entity in the context, for “when”, the first date, for “where”, the first location, and for “what”, the first organization, event, or work of art. The results are in the First Word NER column of Table \ref{tab:ner}. With the exception of NewsQA, we are able to achieve over 40\% of baseline performance with this NER system. 

\begin{table}[htb]
\begin{center}
\begin{tabular}{l c c c c}
\toprule
& & First Word & Person\\
Dataset & Baseline & NER & NER\\
\midrule
SQuAD & 75.6 & 30.0 & 26.7\\
TriviaQA & 58.7 & 25.2 & 8.9\\
NQ & 73.5 & 35.9 & 24.1\\
QuAC & 33.3 & 17.2 & 6.0\\
NewsQA & 60.1 & 11.3 & 8.8\\
\toprule
\end{tabular}
\end{center}
\caption{NER baselines on QA datasets}
\label{tab:ner}
\end{table}

Our Person NER baseline returns the first person entity found in each context. The results are shown in Table \ref{tab:ner}. Both NQ and SQuAD achieve 33-35\% of their baseline by only extracting the first person entity. TriviaQA sees a much larger drop when using only person entities, suggesting there is more entity type variety in the TriviaQA test set. These results show that some questions can be answered by extracting entity types and without needing most or all of the question text.

\subsection{Can QA models handle variations in questions?}
The previous section found that models can perform well on test sets even as seemingly important features are stripped from datasets. This section considers the opposite problem: Can models remain robust as features are added to datasets? To analyze this, we run two experiments where we add filler words and negation to test set questions.

\subsubsection{Filler Words}

\begin{table}[hbt]
\begin{center}
\begin{tabular}{l c c}
\toprule
Dataset & Baseline & Filler Words \\
\midrule
SQuAD & 75.6 & 69.5 \\
TriviaQA & 58.7 & 56.5 \\
NQ & 73.5 & 67.6 \\
QuAC & 33.3 & 31.2 \\
NewsQA & 60.1 & 54.8 \\
\bottomrule
\end{tabular}
\end{center}
\caption{F1 scores slightly decrease on test sets where a filler word is added to the question.}
\label{tab:filler}
\end{table}

If a QA model understands a question, it should handle semantically equivalent questions equally well. While previous works have shown poor model performance on paraphrased questions \cite{ribeiro-etal-2018-semantically, gan-ng-2019-improving}, we take an even simpler approach by adding filler words that do not affect the rest of the question. For each question in the test set, we randomly added one of three filler words (\textit{really}, \textit{definitely}, or \textit{actually}) before the main verb, as identified by spaCy. An example is shown in the seventh row of Table~\ref{tab:examples}.

Table~\ref{tab:filler} shows the results of models fine-tuned on their original training sets and evaluated on the filler word test sets. All models drop between 2 to 5 F1 points. Although these drops may seem small, they do show that even a naive approach can hurt performance. It is no surprise that more sophisticated paraphrases of questions cause models to fail. The SQuAD model in particular had better performance when 50\% of the training set was randomly labeled (73.9) than when filler words were added to the test set (69.5), suggesting that our models have become robust to less consequential features.

\subsubsection{Negation}

Negation is an important grammatical construction for QA systems to understand. Giving the same answer to a question and its negative (Who invented the telescope? vs. Who didn’t invent the telescope?) can frustrate or mislead users. In previous work, \citet{kassner2019negated} studied negation by manually negating 305 SQuAD 1.1 questions and found that a BERT model largely ignored negation. We expand on this work by using the full SQuAD 2.0 dataset and comparing performance across five datasets. 

\begin{table}[hbt]
\begin{center}
\begin{tabular}{l c c c}
\toprule
Dataset & Baseline & Negation \\
\midrule
SQuAD & 75.6  & 2.0\\
TriviaQA & 58.7 & 42.0\\
NQ & 73.5 & 68.9 \\
QuAC & 33.3 & 16.1 \\
NewsQA & 60.1 & 52.3 \\
\bottomrule
\end{tabular}
\end{center}
\caption{With the exception of SQuAD, models continue to return the original answer when given a negated question.}
\label{tab:negation}
\end{table}

For each dataset, we negated every question in the test set by mapping common verbs (i.e. \textit{is}, \textit{did}, \textit{has}) to their contracted negative form (i.e. \textit{isn't}, \textit{didn't}, \textit{hasn't}) or by inserting \textit{never} before the main verb, as identified by spaCy. We keep the original answers in the test set since we want to evaluate how often the original answer is returned for the negative question. In this case, a lower F1 score means better performance. An example is in the last row of Table~\ref{tab:examples}.

We used the models fine-tuned on their original training sets and evaluated them on the negated test sets. The results are in Table~\ref{tab:negation} and show how often each model continued to return the original answer given a negative question. We see that SQuAD outperforms all the other models by giving the original answer to a negative question less than 3\% of the time. Other models return the original answer to the negative question between 48\% and 94\% of the time, suggesting that the negation is largely ignored.

\begin{table}[hbt]
\begin{center}
\begin{tabular}{l c c}
\toprule
Dataset & \textit{n't} & \textit{never} \\
\midrule
SQuAD & 0.85 & 0.89\\
TriviaQA & 0.31 & 0.48\\
NQ & 0.37 & 0.34\\
QuAC & 0.17 & 0.17\\
NewsQA & 0.14 & 0.06\\
\bottomrule
\end{tabular}
\end{center}
\caption{The percentage of questions in the training set containing \textit{n't} or \textit{never} that are impossible}
\label{tab:neg-fraction}
\end{table}

Does the SQuAD model understand negation, or is this a sign of bias? Table~\ref{tab:neg-fraction} shows how often a question containing \textit{n’t} or \textit{never} was impossible in the training set. SQuAD has a high bias, with 85\% of questions containing \textit{n't} and 89\% of questions containing \textit{never} being impossible. This difference could be a result of SQuAD annotators having a bias to include \textit{n't} or \textit{never} more often in impossible questions than answerable questions, while impossible questions in other datasets were more organically collected. This suggests that SQuAD’s performance is due to an annotation artifact. These results find that no dataset adequately prepares a model to understand negation.

\section{Conclusions}
In this work, we probed five QA datasets with six tasks and found that our models did not learn to generalize well, remained suspiciously robust to incorrect or missing data, and failed to handle variations in questions. These findings show that models learn simple heuristics around question-context overlap or entity types and pick up on underlying patterns in the datasets that allow them to remain robust to corrupt examples but not to valid variations. The shortcomings in datasets and evaluation methods make it difficult to judge if models are learning reading comprehension. Based on our work, we make the following recommendations to researchers who create or evaluate QA datasets:

\begin{itemize}
\setlength\itemsep{0.1em}
\item \textbf{Test for generalizability}: Models are more valuable to real-world applications if they generalize. New QA model should report performance across multiple relevant datasets.
\item \textbf{Challenge the models}: Evaluating on too many easy questions can inflate our judgement of what a model has learned. Discard questions that can be solved trivially by high overlap or extracting the first named entity.
\item \textbf{Be wary of cheating}: Good performance does not mean good understanding. Probe datasets by adding noise, shuffling contexts, or providing incomplete input to ensure models aren’t taking shortcuts.
\item \textbf{Include variations}: Models should be prepared to handle a variety of questions. Include variations such as filler words or negation to existing questions to evaluate how well models have understood a question.
\item \textbf{Standardize dataset formats}: When creating new datasets, consider following a standardized format, such as SQuAD, to make cross-dataset evaluations simpler.
\end{itemize}

\bibliography{emnlp2020}
\bibliographystyle{acl_natbib}

\end{document}